\documentclass{IEEEtran}
\usepackage[pdftex]{graphicx}
\usepackage[T1]{fontenc}
\usepackage[utf8]{inputenc}
\usepackage[numbers]{natbib}
\usepackage{times}
\usepackage{latexsym}
\usepackage{MnSymbol}
\usepackage{amsmath}
\usepackage[capitalize]{cleveref}
\usepackage{amsfonts}
\usepackage{url}
\usepackage{color,soul}
\usepackage{bm}
\usepackage{multirow}
\usepackage[font=small]{caption}
\usepackage{subcaption}
\usepackage[noend]{algpseudocode}
\usepackage{afterpage}
\usepackage{tabularx}
\usepackage{booktabs}
\usepackage{arydshln}
\usepackage[shortlabels]{enumitem}
\usepackage{bigstrut}
\usepackage{array}
\usepackage{pifont}
\usepackage{textcomp}
\def\BibTeX{{\rm B\kern-.05em{\sc i\kern-.025em b}\kern-.08em
		T\kern-.1667em\lower.7ex\hbox{E}\kern-.125emX}}

\usepackage{floatrow}
\setkeys{Gin}{width=\linewidth}

\definecolor{mypink1}{rgb}{0.858, 0.188, 0.478}
\definecolor{maroon}{RGB}{220, 20, 60}
\definecolor{green}{RGB}{0, 100, 0}

\definecolor{mypink1}{rgb}{0.858, 0.188, 0.478}
\definecolor{maroon}{RGB}{220, 20, 60}

\begin{document}
	\title{Emotion Recognition in Conversation: Research Challenges, Datasets, and Recent Advances}
	
	\author{Soujanya Poria$^{1}$, Navonil Majumder$^2$, Rada Mihalcea$^3$,
		Eduard Hovy$^4$ \\
		\vspace{5mm}
		\small{ $^1$ School of Computer Science and Engineering, NTU, Singapore} \\
		\small{$^2$ CIC, Instituto Polit\'ecnico Nacional, Mexico} \\
		\small{$^3$ Computer Science \& Engineering, University of Michigan, USA}\\
		\small{ $^4$ Language Technologies Institute, Carnegie Mellon University, USA}\\ 
		\tt sporia@ntu.edu.sg, \tt navo@nlp.cic.ipn.mx, \\
		\tt mihalcea@umich.edu, \tt hovy@cs.cmu.edu  \\
	}
	\maketitle
	
	
	\begin{abstract}
		Emotion is intrinsic to humans and consequently emotion understanding is a key part of human-like
		artificial intelligence (AI).
		Emotion recognition in conversation (ERC) is  becoming increasingly popular as a new research frontier in natural language
		processing (NLP) due to its ability to mine opinions from the plethora of publicly available conversational data in platforms
		such as Facebook, Youtube, Reddit, Twitter, and others. Moreover, it has potential applications in health-care systems (as a
		tool for psychological analysis), education (understanding student frustration) and more. Additionally, ERC is also extremely important for generating emotion-aware dialogues that require an understanding of the user's emotions. Catering
		to these needs calls for effective and scalable conversational emotion-recognition algorithms. However, it is a strenuous problem to solve because of several research challenges. In this paper, we discuss these challenges and shed light on the recent research in this  field. We also describe the drawbacks of these approaches and discuss the reasons why they fail to successfully overcome the research challenges in ERC.
	\end{abstract}
	
	\section{Introduction}
	
	Emotion is often defined as an individual's mental state associated with thoughts, feelings
	and behaviour. 
	Stoics like Cicero organized emotions into four categories - {\it metus} (fear), {\it aegritudo} (pain), {\it libido} (lust) and {\it laetitia} (pleasure).
	Later, evolutionary theory of emotion were initiated in the late 19th century by Charles Darwin~\citep{darwin1998expression}. He hypothesized that emotions evolved through natural selection and, hence, have cross-culturally universal counterparts. In recent times, \citet{plutchik}
	categorized emotion into eight primary types, visualized by the wheel of
	emotions (\cref{fig:plutchik}). Further, \citet{ekman1993facial} argued the correlation
	between emotion and facial expression. 
	
	Natural language is
	often indicative of one's emotion. Hence, emotion recognition has
	been enjoying popularity in the field of NLP  \citep{kratzwald2018decision, colneric2018emotion}, due to its widespread applications in
	opinion mining, recommender systems, health-care, and so on. \citet{Strapparava:2007:STA:1621474.1621487} addressed the task of  emotion detection on news headlines. A number of emotion lexicons~\citep{strapparava2004wordnet,mohammad2010emotions} have been developed to tackle the textual emotion recognition problem. 
	
	Only in the past few years has emotion recognition in conversation (ERC) gained attention from the NLP community~\citep{yeh2019interaction, chen2018emotionlines,majumder2019dialoguernn,zhou2018emotional} 
	due to the increase of public availability of conversational data. ERC can be used to analyze
	conversations that take place on social media. It can also aid in analyzing conversations in real times, which can be instrumental in legal trials, interviews, e-health services and more.
	
	Unlike vanilla emotion recognition of sentences/utterances, ERC ideally requires context
	modeling of the individual utterances. This context can be attributed to the preceding
	utterances, and relies on the temporal sequence of utterances.
	Compared to the recently published works on ERC~\citep{chen2018emotionlines,majumder2019dialoguernn,zhou2018emotional}, both lexicon-based~\citep{wu2006emotion,mohammad2010emotions, shaheen2014emotion} and modern deep learning-based~\citep{kratzwald2018decision, colneric2018emotion} vanilla emotion recognition approaches fail to work well on ERC datasets as these works ignore the conversation specific factors such as the presence of contextual cues, the temporality in speakers' turns, or speaker-specific information. \cref{fig:context1} and \cref{fig:context2} show an example where the same
	utterance changes its meaning depending on its preceding utterance. 
	
\begin{figure}[h] 
		\centering 
		\small
		\includegraphics[width=\linewidth]{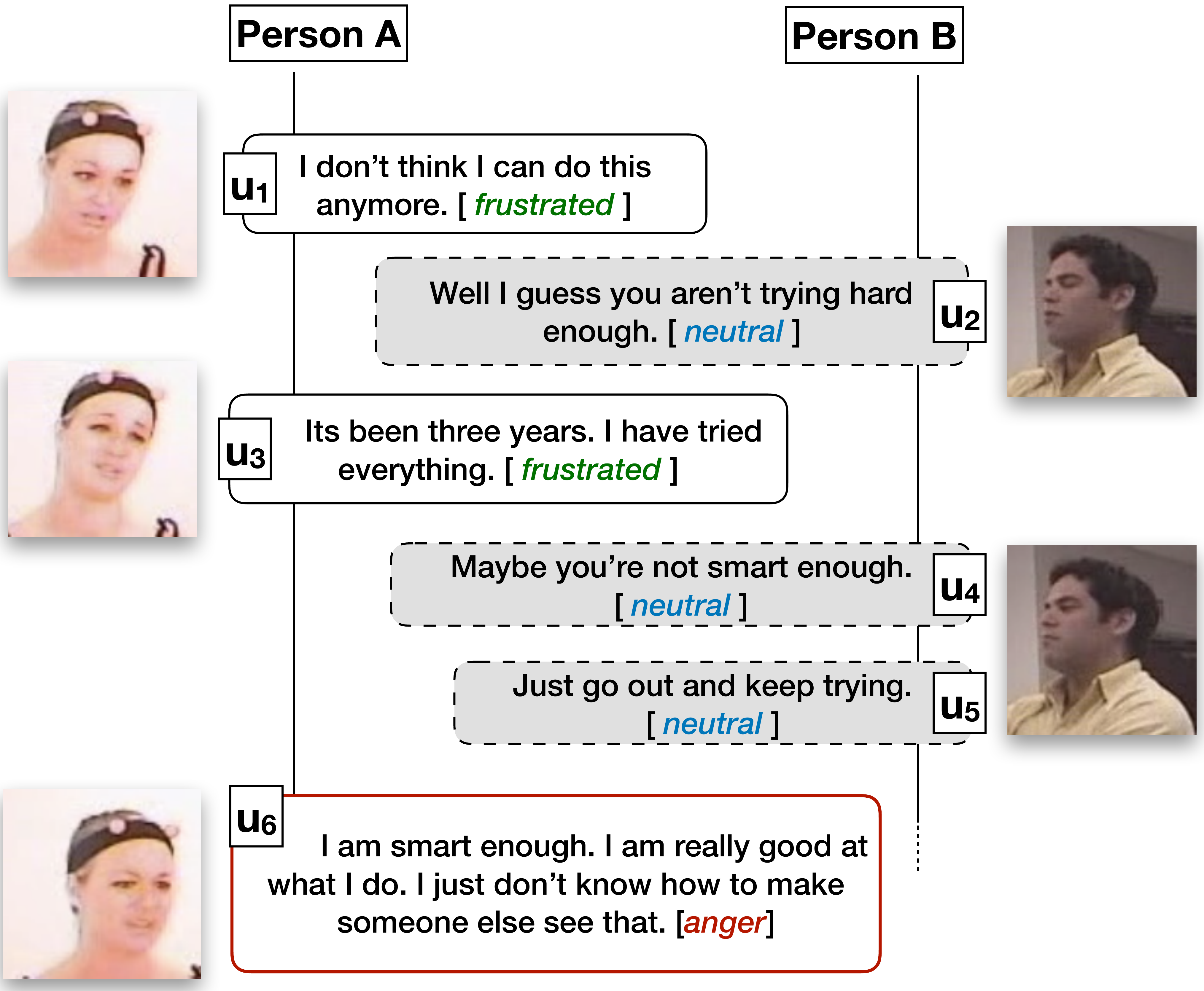} 
		\caption{An abridged dialogue from the IEMOCAP dataset.}
		\label{fig:example}
	\end{figure}

	\textbf{Task definition --- }Given the transcript of a conversation along with speaker information
	of each constituent utterance, the ERC task aims to identify the emotion of each utterance from 
	several pre-defined emotions. \cref{fig:example} illustrates one such
	conversation between two people, where each utterance is labeled by the
	underlying emotion. Formally, given the input sequence of $N$ number of
	utterances $[(u_1, p_1), (u_2,p_2),\dots, (u_N,p_N)]$, where each utterance $u_i=[u_{i,1},u_{i,2},\dots,u_{i,T}]$ consists of $T$ words $u_{i,j}$ and spoken by
	party $p_i$, the task is to predict the emotion label $e_i$ of
	each utterance $u_i$.
	
\begin{figure}
		\centering
		\includegraphics{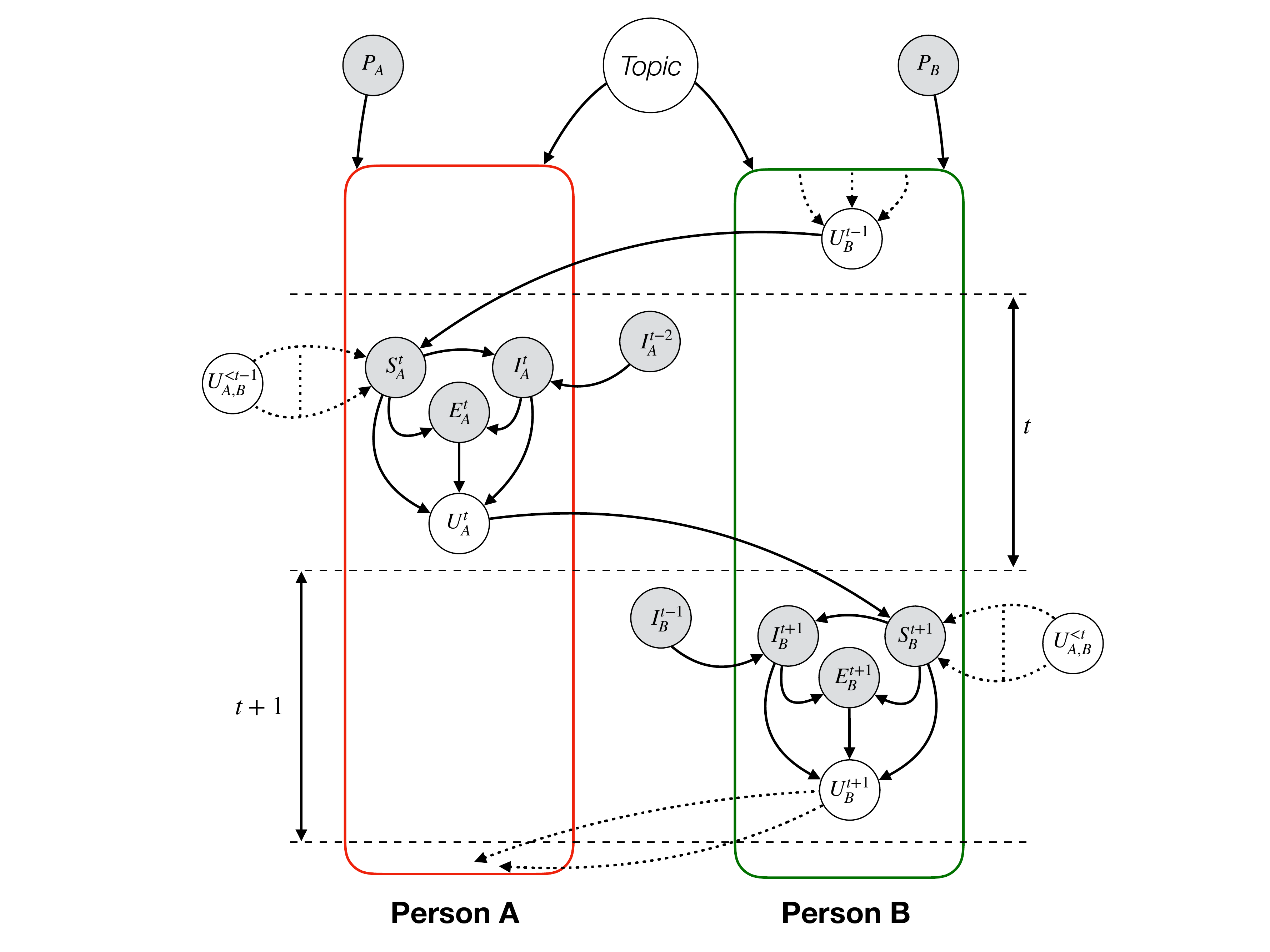}
		\caption{Interaction among different variables during a dyadic conversation between persons A and B. Grey and white circles represent hidden and observed variables, respectively. $P$ represents personality, $U$ represents utterance, $S$ represents interlocutor state, $I$ represents interlocutor intent, $E$ represents emotion and \emph{Topic} represents topic of the conversation. This can easily be extended to multi-party conversations.}
		\label{fig:intent-modelling}
	\end{figure}

	\textbf{Controlling variables in conversations ---} Conversations are broadly categorized into two categories: task oriented and chit-chat (also called as non-task oriented). Both kinds of conversation are governed by different factors or pragmatics~\cite{hovy1987generating}, such as topic, interlocutors' personality, argumentation logic, viewpoint, intent~\citep{schloder2015clarifying}, and so on. \cref{fig:intent-modelling} depicts how these factors play out in a dyadic conversation. Firstly, topic ($Topic$) and interlocutor personality ($P_*$) always influence the conversation, irrespective of the time. A speaker makes up his/her mind ($S^t_*$) about the reply ($U^t_*$) based on the contextual preceding utterances ($U^{<t}_*$) from both speaker and listener, the previous utterance being the most important one since it usually makes the largest change in the joint task model (for task-oriented conversations) or the speaker's emotional state (for chit-chat). Delving deeper, the pragmatic features, as explained by \citet{hovy1987generating}, like argumentation logic, interlocutor viewpoint, inter-personal relationship and dependency, situational awareness are encoded in speaker state ($S^t_*$). Intent ($I^t_*$) of the speaker is decided based on previous intent $I_*^{t-2}$ and speaker state $S_*^t$, as the interlocutor may change his/her intent based on the opponent's utterance and current situation. Then, the speaker formulates appropriate emotion $E_*^t$ for the response based on the state $S^t_*$ and intent $I^t_*$. Finally, the response $U^t_*$ is produced based on the speaker state $S^t_*$, intent $I^t_*$ and emotion $E^t_*$. We surmise that considering these factors would help representing the argument and discourse structure of the conversation, which leads to improved conversation understanding, including emotion recognition. 
	
	Early computational work on dialogue focused mostly on task-oriented cases, in which the overall conversational intent and step-by-step sub-goals played a large part~\citep{grosz1986attention,appelt1992planning}. \citet{cohen1985speech} developed a model and logic to represent intentions and their connections to utterances, whose operators explicate the treatment of beliefs about the interlocutor’s beliefs and vice versa, recursively.  Emotion however played no role in this line of research.  In more recent work, chatbots and chit-chat dialogue have become more prominent, in part due to the use of distributed (such as embedding) representations that do not readily support logical inference.
	
\begin{figure}
		\centering
		\includegraphics[width=\linewidth]{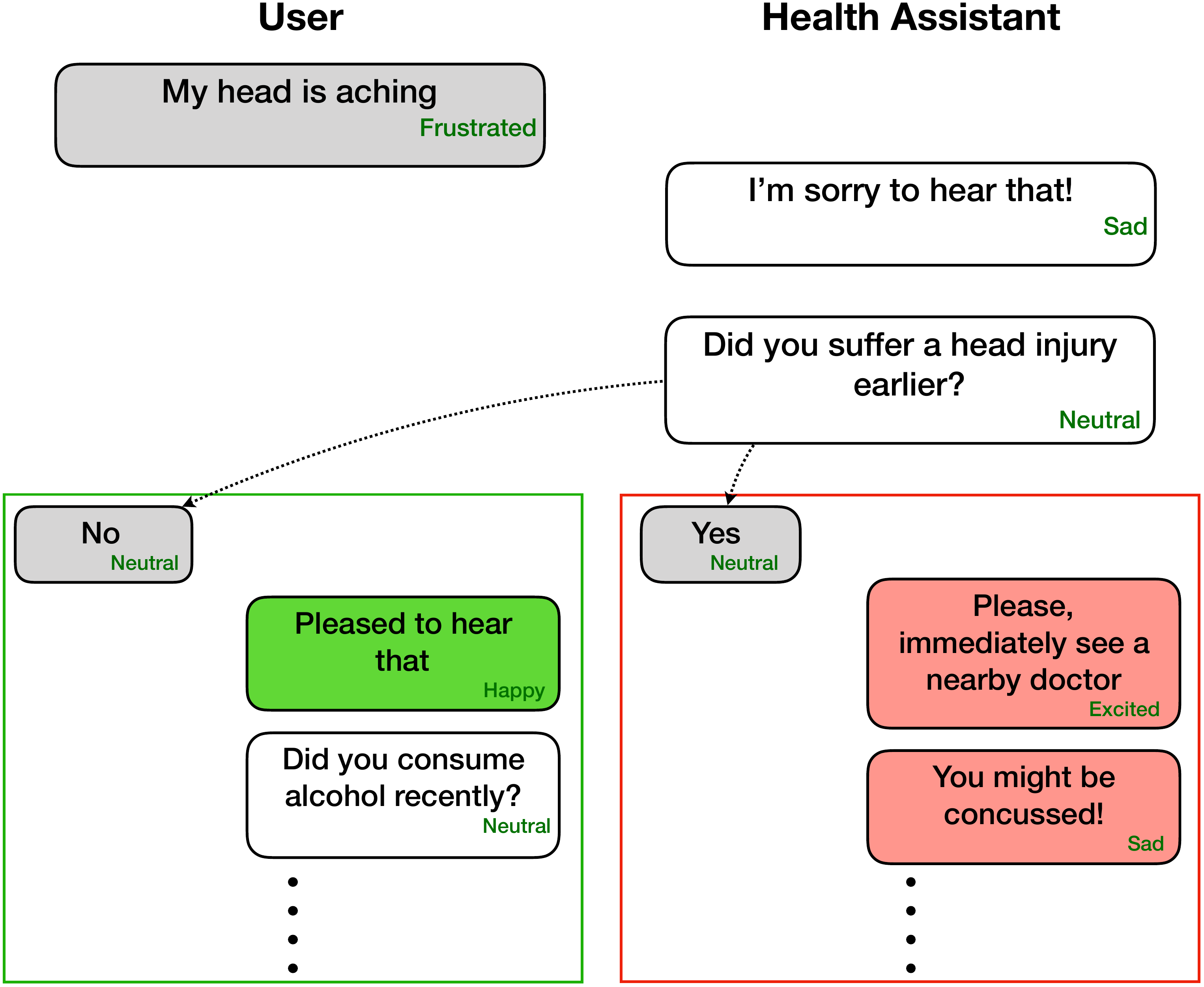}
		\caption{Illustration of an affective conversation where the emotion depends on the context.}
		\label{fig:affective-dialogue}
	\end{figure}

	On conversational setting, \citet{dmello} and \citet{yang} worked with small datasets with three and four emotion labels, respectively. This was followed by \citet{Y16-2006}, where emotion detection on conversation transcript was attempted. Recently, several
	works~\citep{hazarika2018icon,bae2019snu_ids} have devised deep learning-based techniques for ERC.
	These works are crucial as we surmise an instrumental role
	of ERC in emotion-aware a.k.a. affective dialogue generation which has fallen within the topic of ``text generation under pragmatics constriants'' as proposed by~\citet{hovy1987generating}. \cref{fig:affective-dialogue} illustrates
	one such conversation between a human ({\it user}) and a medical chatbot ({\it health-assistant}). The assistant responds with emotion based
	on the {\it user}'s input. Depending on whether the {\it user} suffered an injury earlier or not, the
	{\it health-assistant} responds with excitement (evoking urgency) or happiness (evoking relief).
	
	As ERC is a new research field, outlining research challenges, available datasets, and benchmarks can potentially aid future research on ERC. In this paper, we aim to serve this purpose by discussing various factors that contribute to the emotion dynamics in a conversation. We surmise that this paper will not only help the researchers to better understand the challenges and recent works on ERC but also show possible future research directions. The rest of the paper is organized as follows: \cref{sec:challenge} presents the key research challenges; \cref{sec:datasets} and \ref{sec:benchmarks} cover the datasets and recent progress in this field; finally \cref{sec:conclusion} concludes the paper. 
	
	\section{Research Challenges}
	\label{sec:challenge}
	
	Recent works on ERC, e.g., DialogueRNN \cite{majumder2019dialoguernn} or ICON \cite{hazarika2018icon}, strive to address several key research challenges that make the
	task of ERC difficult to solve:
	
	\paragraph{Categorization of emotions}
	
	Emotion is defined using two type of models --- categorical and
	dimensional. Categorical model classifies emotion into a fixed number of
	discrete categories. In contrast, dimensional model describes emotion as a point in
	a continuous multi-dimensional space.
	
\begin{figure}
		\centering
		\includegraphics[width=\linewidth]{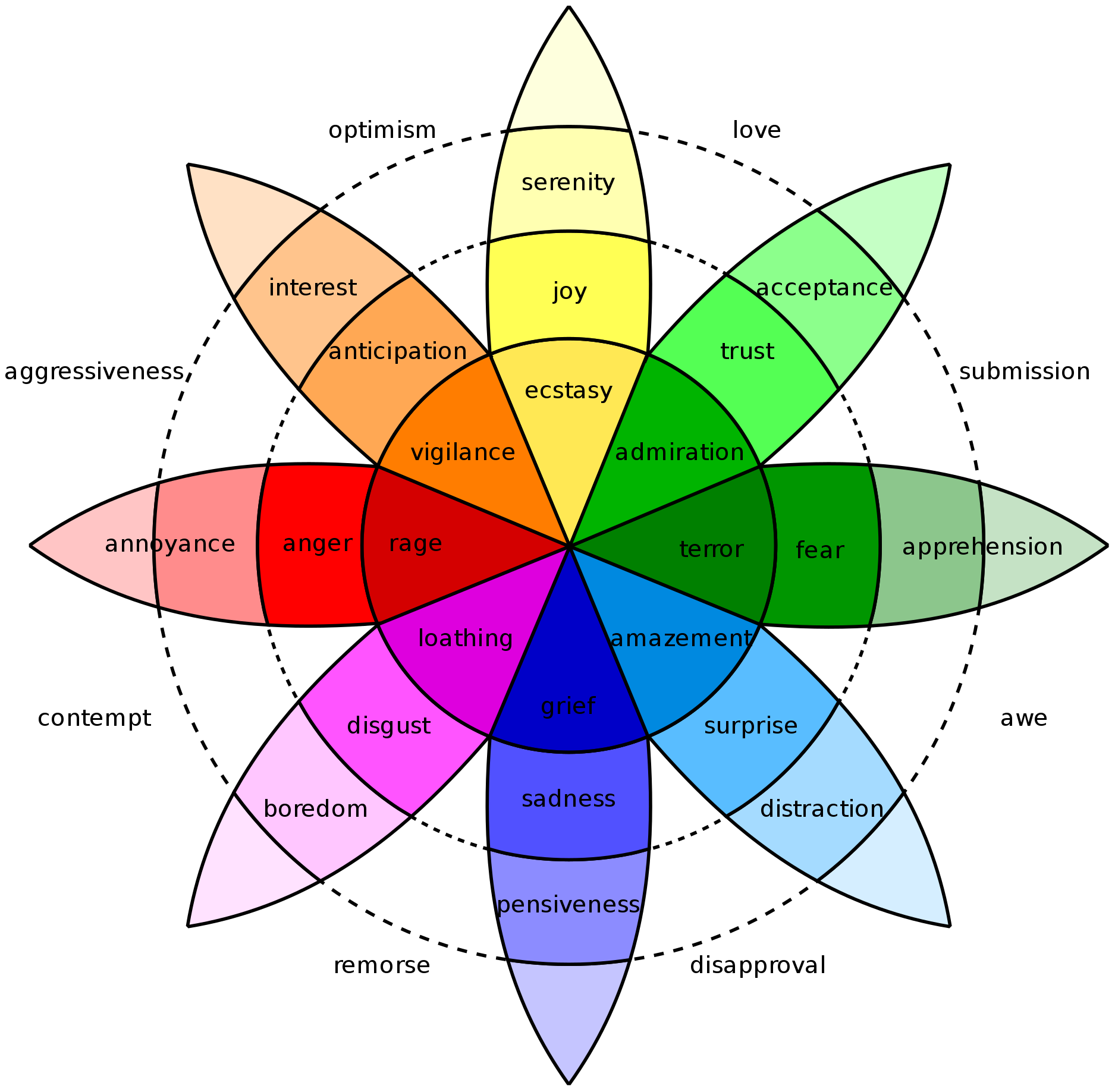}
		\caption{Plutchik's wheel of emotion~\citep{plutchik}.}
		\label{fig:plutchik}
	\end{figure}

	In the categorical front, \citet{plutchik}'s wheel of emotions~(\cref{fig:plutchik}) defines eight
	discrete primary emotion types, each of which has finer related subtypes. 
	On the other hand, \citet{ekman1993facial} concludes six basic emotions ---
	anger, disgust, fear, happiness, sadness and surprise.
	
	Most dimensional categorization models~\citep{jamesrussel,Mehrabian1996} adopt two dimensions
	--- valence and arousal. Valence represents the degree of emotional positivity, and arousal represents the intensity of the emotion. In contrast with
	the categorical models, dimensional
	models map emotion into a continuous spectrum rather than hard categories. This enables easy and intuitive comparison of two emotional states using vector operations, whereas comparison is non-trivial for categorical models. As there are multiple categorization and dimensional taxonomies available, it is challenging to select one particular model for annotation. Choosing a simple categorization model e.g., Ekman's model has a major drawback as these models are unable to ground complex emotions. On the other hand, complex emotion models such as Plutchik's model make it very difficult for the annotators to discriminate between the related emotions, e.g., discerning anger from rage. Complex emotion models also increase the risk of obtaining a lower inter-annotator agreement.
	
	The popular ERC dataset IEMOCAP~\citep{busso2008iemocap} adopted both
	categorical and dimensional models. However, newer ERC datasets like
	DailyDialogue~\citep{li2017dailydialog} have employed only categorical model due to its more 
	intuitive nature. Most of the available datasets for emotion recognition in conversation adopted simple taxonomies, which are slight variants of Ekman's model. Each emotional utterance in the EmoContext dataset is labeled with one of the following emotions: \emph{happiness, sadness and anger}. The majority of the utterances in EmoContext do not elicit any of these three emotions and are annotated with an extra label: \emph{others}. Naturally, the inter-annotator agreement for the EmoContext dataset is higher due to its simplistic emotion taxonomy. However, the short context length and simple emotion taxonomy make ERC on this dataset less challenging.
	
	\paragraph{Basis of emotion annotation} 
	
	Annotation with emotion labels is challenging as the label depends on the annotators
	perspective. Self-assessment by the interlocutors in a conversation is arguably
	the best
	way to annotate utterances. However, in practice it is unfeasible as real-time 
	tagging of unscripted conversations will impact the conversation flow.
	Post-conversation self-annotation could be an option, but it has not been done yet.
	
	As such, many ERC datasets~\citep{busso2008iemocap} are scripted and annotated by a
	group of people
	uninvolved with the script and conversation. The annotators are given the context
	of the utterances as prior knowledge for accurate annotation. Often pre-existing transcripts are annotated for quick turn-around, as in EmotionLines~\citep{chen2018emotionlines}. 
	
	The annotators also need to be aware of the interlocutors
	perspective for situation-aware annotation. For example, the emotion behind the
	utterance ``\textit{Lehman Brothers' stock is plummeting!!}'' depends on whether
	the speaker benefits from the crash. The annotators should be aware of the nature
	of association between the speaker and Lehman Brothers for accurate labeling.
	
	\paragraph{Conversational context modeling}
	
	Context is at the core of the NLP research. According to several recent studies~\citep{peters2018deep,devlin2018bert}, contextual sentence and word embeddings can improve the performance of the state-of-the-art NLP systems by a significant margin.
	
	The notion of context can vary from problem to problem. For example, while calculating word representations, the surrounding words carry contextual information. Likewise, to classify a sentence in a document, other neighboring sentences are considered as its context. In \citet{poria2017context}, surrounding utterances are treated as context and they experimentally show that contextual evidence indeed aids in classification.
	
	Similarly in conversational emotion-detection, to determine the emotion of an utterance at time $t$, the preceding utterances at time $<t$ can be considered as its context. However, computing this context representation often exhibits major difficulties due to emotional dynamics.
	
	Emotional dynamics of conversations consists of two important aspects: \textit{self} and \textit{inter-personal dependencies}~\citep{morris2000emotions}. self-dependency, also known as \textit{emotional inertia}, deals with the aspect of emotional influence that speakers have on themselves during conversations~\citep{kuppens2010emotional}. 
	On the other hand, inter-personal dependencies relate to the emotional influences that the counterparts induce into a speaker. Conversely, during the course of a dialogue, speakers also tend to mirror their counterparts to build rapport~\citep{navarretta2016mirroring}. This phenomenon is illustrated in \cref{fig:example}. Here, $P_a$ is frustrated over her long term unemployment and seeks encouragement ($u_1, u_3$). $P_b$, however, is pre-occupied and replies sarcastically ($u_4$). This enrages $P_a$ to appropriate an angry response ($u_6$). In this dialogue, \textit{emotional inertia} is evident in $P_b$ who does not deviate from his nonchalant behavior. $P_a$, however, gets emotionally influenced by $P_b$. Modeling self and inter-personal relationship and dependencies may also depend on the topic of the conversation as well as various other factors like argument structure, interlocutors’ personality, intents, viewpoints on the conversation, attitude towards each other etc.. Hence, analyzing all these factors are key for a true self and inter-personal dependency modeling that can lead to enriched context understanding. 
	
	The contextual information can come from both local and distant conversational history. While the importance of local context is more obvious, as stated in  recent works, distant context often plays a less important role in ERC. Distant contextual information is useful mostly in the scenarios when a speaker refers to earlier utterances spoken by any of the speakers in the conversational history.
	
\begin{figure*}[!htbp]
		\centering
		\begin{subfigure}{0.49\textwidth}
			\centering
			\includegraphics[width=\linewidth]{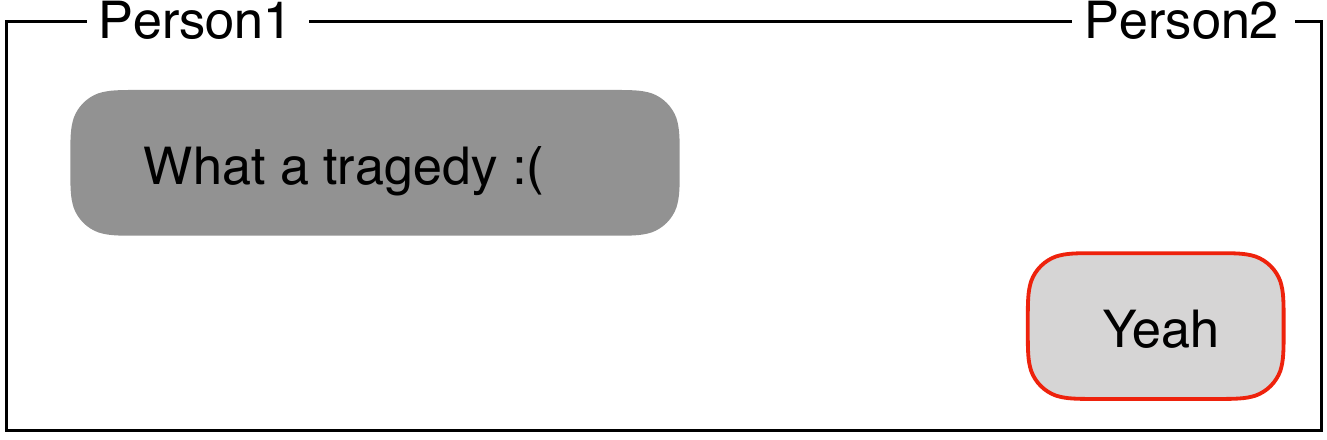}
			\caption{Negative emotion expressed by the utterance \emph{Yeah}.}
			\label{fig:context1}
		\end{subfigure}
		\begin{subfigure}{0.49\textwidth}
			\centering
			\includegraphics[width=\linewidth]{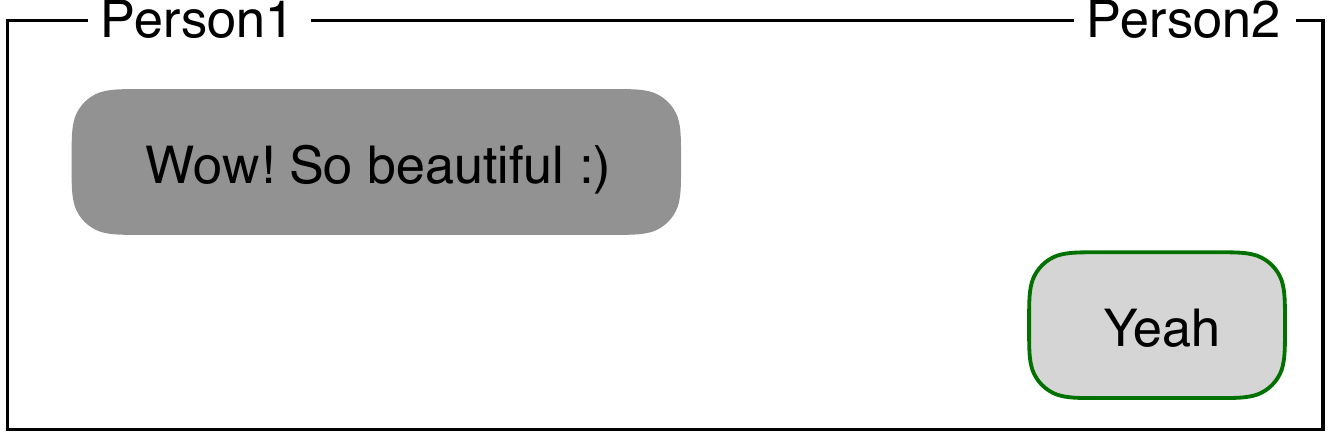}
			\caption{Positive emotion expressed by the utterance \emph{Yeah}.}
			\label{fig:context2}
		\end{subfigure}
		\caption{Role of context in emotion recognition in conversation.}
	\end{figure*}

	The usefulness of context is more prevalent in classifying short utterances, like {\it ``yeah''}, {\it ``okay''}, {\it ``no''}, that
	can express different emotions depending on the context and discourse of the dialogue. The examples in \cref{fig:context1} and \cref{fig:context2} explain this phenomenon. The emotions expressed by the same utterance ``\emph{Yeah}'' in both these examples differ from each other and can only be inferred from the context. 
	
	Finding contextualized conversational utterance representations is an active area of research. Leveraging such contextual clues is a difficult task. Memory networks, RNNs, and attention mechanisms have been used in previous works e.g., HRLCE or DialogueRNN, to grasp information from the context.
	
	\paragraph{Speaker specific modeling}
	
	Individuals have their own subtle way of expressing emotions. For instance, some individuals are more sarcastic than
	others. For such cases, the usage of certain words would vary depending on if they are being sarcastic. 
	Let's consider this example, $P_a:$ \textit{``The order has been cancelled.''}, $P_b:$ \textit{``This is great!''}. If $P_b$ is
	a sarcastic person, then his response would express negative emotion to the order being canceled through the word \textit{great}. On the other hand, $P_b$'s
	response, \textit{great}, could be taken literally if the canceled order is beneficial to $P_b$ (perhaps $P_b$ cannot afford the product he ordered). Since, necessary background
	information is often missing from the conversations, speaker profiling based on preceding utterances often yields improved
	results.
	
	\paragraph{Listener specific modeling}
	
	During a conversation, the listeners make up their mind about the speaker's utterance as it's
	spoken. However, there is no textual data on the listener's reaction to the speaker. A model
	must resort to visual modality to model the listener's facial expression to capture the
	listener's reaction. However, according to DialogueRNN, capturing listener
	reaction does not yield any improvement as the listener's subsequent utterance carries their
	reaction. Moreover, of the listener never speaks in a conversation, his/her reaction remains
	irrelevant. Nonetheless, listener modeling can be useful in the scenarios where continuous
	emotion recognition of every moment of the conversation is necessary, like audience reaction
	during a political speech, as opposed to emotion recognition of each utterance.
	
	\paragraph{Presence of emotion shift}
	
\begin{figure*}[h] 
		\centering 
		\small
		\includegraphics[width=0.9\linewidth]{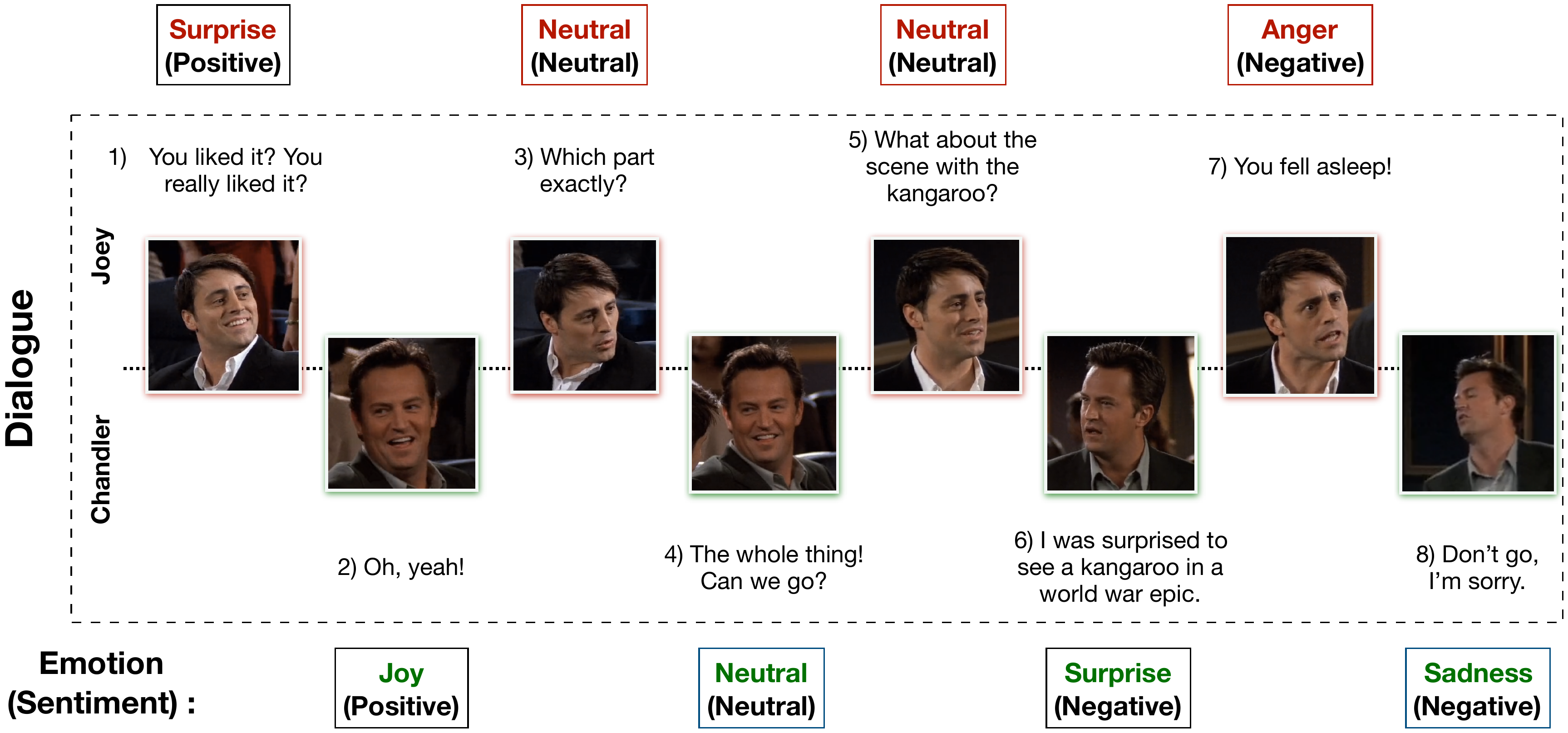}
		\caption{ 
			Emotion shift of speakers in a dialogue in comparison with speaker’s previous emotion. Red
			and blue colors are used to show the emotion shift of Joey and Chandler respectively..}
		\label{fig:exampleshift0}
	\end{figure*}

	Due to \textit{emotional inertia}, participants in a conversation tend to stick a particular emotional state, unless
	some external stimuli, usually the other participants, invoke a change. This is illustrated in \cref{fig:exampleshift0}, where Joey
	changes his emotion from \textit{neutral} to \textit{anger} due to the last utterance of Chandler, which was unexpected and rather shocking to Joey. This is a hard problem
	to solve, as the state-of-the-art ERC model, DialogueRNN is more accurate in emotion detection for the utterances without emotional shift or when the shift is to a similar emotion (e.g., from fear to sad). It is equally interesting and challenging to track the development, shift and dynamics of the participants' opinions on the topic of discussion in the conversation.
	
\begin{figure}[h] 
		\centering 
		\small
		\includegraphics[width=0.8\linewidth]{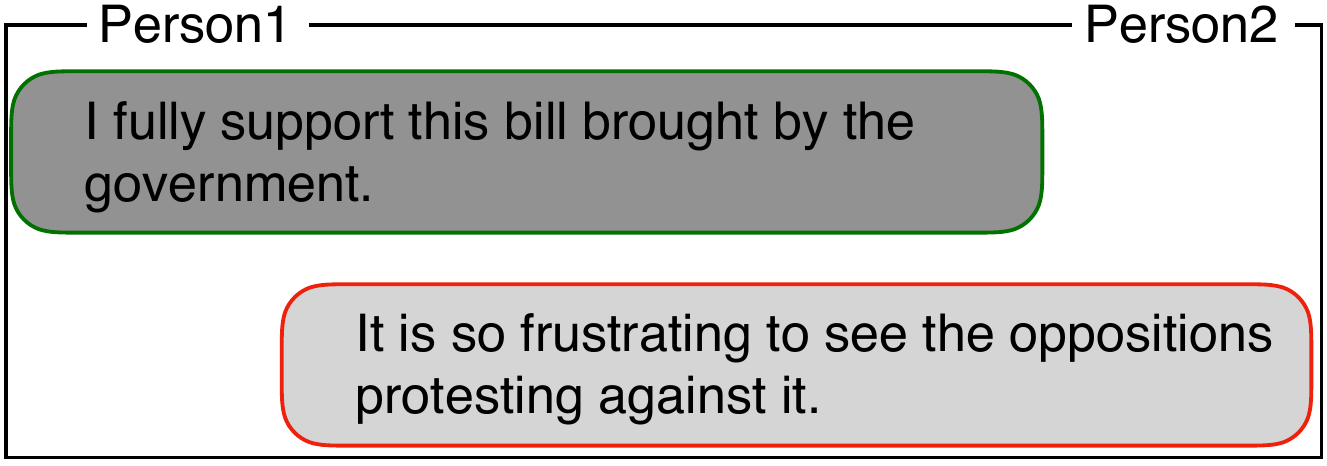} 
		\caption{Fine-grained emotion understanding: an example.}
		\label{fig:exampleshift}
	\end{figure}

	\paragraph{Fine-grained emotion recognition}
	Fine-grained emotion recognition aims at recognizing emotion expressed on the explicit and implicit topics. It involves a deeper understanding of the topic of the conversation, interlocutor opinion and stand. For example, in 
	\cref{fig:exampleshift}, while both persons take a supportive stand for the government's bill, they use completely opposite emotions to express it. It is not possible for a vanilla emotion recognizer to understand the positive emotion of both the interlocutors on the aspect of \emph{government's bill}. Only by interpreting {\it Person 2}'s frustration about the opposition's protest against the bill can a classifier infer {\it Person 2}'s support for the bill. On the other hand, even though {\it Person 1} does not explicitly express his/her opinion on the opposition, from the discourse of the conversation, it can be inferred that {\it Person 1} holds a negative opinion on the opposition.
	
	\paragraph{Multiparty conversation}
	In a multiparty conversation, more than two participants are involved. Naturally, emotion recognition in such conversations is more challenging in comparison with dyadic conversations due to the difficulty in tracking individual speaker states and handling co-references.
	
	\paragraph{Presence of sarcasm}
	
	%
	Sarcasm is a linguistic tool that uses irony to express contempt. An ERC system incapable of detecting sarcasm mostly fails to predict emotion of the sarcastic utterances correctly. Sarcasm detection in a conversation largely depends on the context and discourse of the conversation. For example, the utterance ``\textit{The part where Obama signed it}'' can only be detected as sarcastic if we look at the previous utterance ``\textit{What part of this would be unconstitutional?}''. Sarcastic nature is also person dependent, which again warrants speaker profiling in the conversation.
	
	
	\paragraph{Emotion reasoning}
	
	The ability to reason is necessary for any explainable AI system. In the context of ERC, it is often desired to understand the cause of an expressed emotion by a speaker. As an example, we can refer to \cref{fig:example}. An ideal ERC system, with the ability of emotion reasoning, should perceive the reason for $Person_A$'s \textit{anger}, expressed in $u_6$ of \cref{fig:example}. It is evident upon observation that this \textit{anger} is caused by the persistent nonchalant behavior of $Person_{B}$. Readers should not conflate emotion reasoning with context modeling, which we discuss earlier in this section. Unlike context modeling, emotion reasoning does not only find the contextual utterances in conversational history that triggers the emotion of an utterance, but also determines the function of those contextual utterances on the target utterance. In \cref{fig:example}, it is the \emph{indifference} of $Person_{B}$, reflected by $u_4$ and $u_5$, that makes $Person_{A}$ angry. Similarly, in \cref{fig:exampleshift0}, Joey expresses \emph{anger} once he ascertains Chandler's \emph{deception} in the previous utterance. It is hard to define a taxonomy or tagset for emotion reasoning. At present, there is no available dataset which contains such rich annotations. Building such dataset would enable future dialogue systems to frame meaningful argumentation logic and discourse structure, taking one step closer to human-like conversation. 
	
	\section{Datasets}
	\label{sec:datasets}
	In the last few years, emotion recognition in conversation has gained major research interest, mainly because of its potential application in dialogue systems to generate emotion-aware and empathetic dialogues~\citep{zhou2018emotional}. The primary goal of ERC task is to label each utterance in the conversation with an emotion label. In this section, we discuss the publicly available ERC datasets as well as the shortcomings of these datasets.
	\begin{table*}[t]
		\centering
		\begin{tabular}{cccccc}
			\toprule
			
			Label & DailyDialog & MELD & EmotionLines & IEMOCAP & EmoContext\\
			\midrule
			Neutral & 85572 & 6436 & 6530 & 1708 & -\\
			Happiness/Joy & 12885 & 2308 & 1710 & 648 & 4669\\
			Surprise & 1823 & 1636 & 1658 & - & -\\
			Sadness & 1150 & 1002 & 498 & 1084 & 5838\\
			Anger & 1022 & 1607 & 772 & 1103 & 5954\\
			Disgust & 353 & 361 & 338 & - & -\\
			Fear & 74 & 358 & 255 & - & -\\
			Frustrated & - & - & - & 1849 & -\\
			Excited & - & - & - & 1041 & -\\
			Other & - & - & - & - & 21960\\
			\bottomrule
		\end{tabular}
		\caption{Label distribution statistics in different Emotion Recognition datasets.}
		\label{tab-label}
	\end{table*}
	There are a few publicly available datasets for ERC - IEMOCAP~\citep{busso2008iemocap}, SEMAINE~\citep{mckeown2012semaine}, Emotionlines~\citep{chen2018emotionlines}, MELD~\citep{poria2018meld}, DailyDialog~\citep{li2017dailydialog} and EmoContext~\citep{chatterjee2019understanding}. A detailed comparison of these datasets is drawn in \cref{table:data}. Out of these five datasets, IEMOCAP, SEMAINE and MELD are multimodal (containing acoustic, visual and textual information) and the remaining two are textual. Apart from SEMAINE dataset, rest of the datasets contains categorical emotion labels. In contrast, each utterance of SEMAINE dataset is annotated with
	four real valued affective attributes: valence ($[-1,1]$), arousal ($[-1,1]$),
	expectancy ($[-1,1]$) and power ($[0,\infty)$).  We also show the emotion label distribution of these datasets in \cref{tab-label}. In EmoContext dataset, an emotion label is assigned to only the last utterance of each dialogue.
	None of these datasets can be used for emotion reasoning as they lack necessary annotation details required for the reasoning task. Readers should also note that, all these datasets do not contain fine-grained and topic level emotion annotation.
	
	\section{Recent advances}
	\label{sec:benchmarks}
	In this section we give a brief introduction to the recent work on this topic. We also compare the approaches and report their drawbacks.
	\begin{table*}[h!]
		\centering
		\begin{tabular}{c|c|c|c|c|c|c}
			\toprule
			\multirow{2}{*}{Dataset}&\multicolumn{3}{c|}{$\#$ dialogues}&\multicolumn{3}{c}{$\#$ utterances}\\
			&train&val&test&train&val&test\\
			\hline \hline
			IEMOCAP&\multicolumn{2}{c|}{120}&31&\multicolumn{2}{c|}{5810}&1623\\
			SEMAINE&\multicolumn{2}{c|}{63}&32&\multicolumn{2}{c|}{4368}&1430\\
			EmotionLines&720&80&200&10561&1178&2764\\
			MELD&1039&114&280& 9989& 1109& 2610\\
			DailyDialog&11,118&1000&100&87,832&7912&7863\\
			EmoContext&30,159&2754&5508&90,477&8262&16,524\\
			\bottomrule
		\end{tabular}
		\caption{Comparison among IEMOCAP, SEMAINE, EmotionLines, MELD and DailyDialog datasets.}
		\label{table:data}
	\end{table*}
	
	As depicted in \cref{fig:intent-modelling}, recognizing emotion of an utterance in a conversation primarily depends on these following three factors: 
	\begin{enumerate}
		\item the utterance itself and its context defined by the interlocutors' preceding utterances in the conversation, as well as intent and the topic of the conversation,
		\item the speaker's state comprising variables like personality and argumentation logic and,
		\item emotions expressed in the preceding utterances.
	\end{enumerate}
	Although, IEMOCAP and SEMAINE have been developed almost a decade ago, most of the works that used these two datasets did not consider the aforementioned factors.
	\paragraph{Benchmarks and their drawbacks}
	Based on these factors, a number of approaches to address the ERC problem have been proposed recently. Conversational memory network (CMN), proposed by  \citet{hazarika2018conversational} for dyadic dialogues, is one of the first ERC approaches that utilizes distinct memories for each speaker for speaker-specific context modeling. Later, \citet{hazarika2018icon} improved upon this approach with interactive conversational memory network (ICON), which interconnects these memories to model self and inter-speaker emotional influence. None of these two methods actually exploit the speaker information of the target utterance for classification. This makes the model blind to speaker-specific nuances. Recently, \citet{yeh2019interaction} proposed an ERC method called Interaction-aware Attention Network (IANN) by leveraging inter-speaker relation modeling. Similar to ICON and CMN, IANN utilises distinct memories for each speaker.  
	\begin{figure*}[h]
		\centering
		\includegraphics[width=0.85\linewidth]{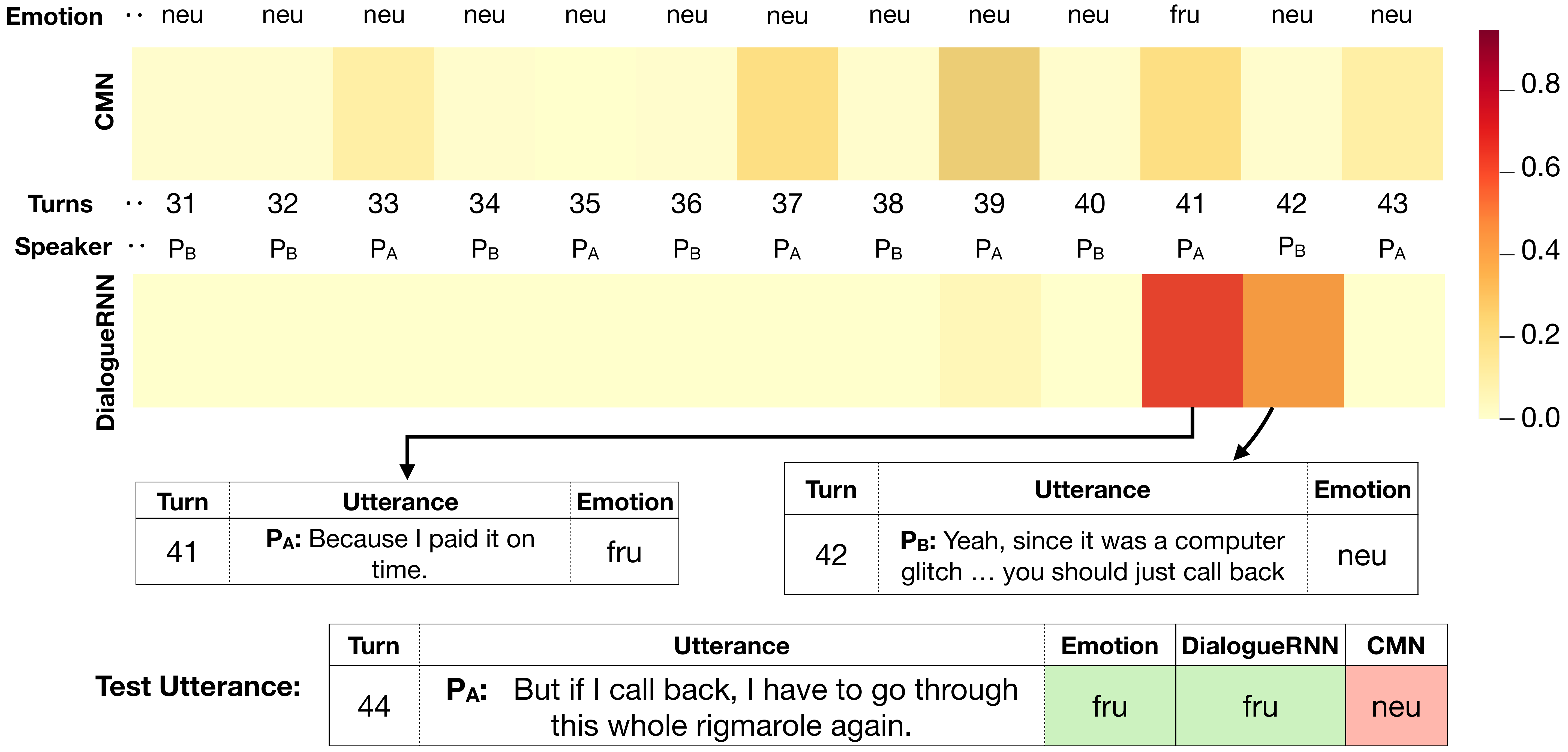}
		\caption{Comparison of attention scores over utterance history of CMN and DialogueRNN. Higher attention value signifies more important contextual information. Note: Figure taken from \citet{majumder2019dialoguernn}.}
		\label{fig:case-study}
	\end{figure*}
	DialogueRNN~\citep{majumder2019dialoguernn} aims to solve this issue by considering the speaker information of the target utterance and, further, modeling self and inter-speaker emotional influence with a hierarchical multi-stage RNN with attention mechanism. On both IEMOCAP and SEMAINE datasets, DialogueRNN outperformed~(\cref{tab:results-text1} and \cref{tab:results-text2}) the other two approaches.
	
	\begin{table*}[ht!]
		\centering
		\resizebox{\linewidth}{!}{
			\begin{tabular}{c||c@{~~}c|c@{~~}c|c@{~~}c|c@{~~}c|c@{~~}c|c@{~~}c|c@{~~}c}
				\hline
				\multirow{3}{*}{Methods} & \multicolumn{14}{c}{IEMOCAP}\\
				\cline{2-15} & \multicolumn{2}{c|}{Happy} & \multicolumn{2}{c|}{Sad} &
				\multicolumn{2}{c|}{Neutral} & \multicolumn{2}{c|}{Angry} & \multicolumn{2}{c|}{Excited} & \multicolumn{2}{c|}{Frustrated} & \multicolumn{2}{c}{\textbf{Average(w)}}\\
				\cline{2-15} & Acc. & F1 & Acc. & F1 & Acc. & F1 & Acc. & F1 & Acc. & F1 & Acc. & F1 & Acc. & F1\\
				\hline
				\hline
				CNN &27.77&29.86&57.14&53.83&34.33&40.14&61.17&52.44&46.15&50.09&62.99&55.75&48.92&48.18 \\
				memnet &25.72&33.53&55.53&61.77&58.12&52.84&59.32&55.39&51.50&58.30&67.20&59.00&55.72&55.10\\
				bc-LSTM &29.17&34.43&57.14&60.87&54.17&51.81&57.06&56.73&51.17&57.95&67.19&58.92&55.21&54.95 \\
				bc-LSTM+Att &30.56&{\bf 35.63}&56.73&62.90&57.55&53.00&59.41&59.24&52.84&58.85&65.88&59.41&56.32&56.19 \\
				\hline
				CMN &25.00&30.38&55.92&62.41&52.86&52.39&61.76&59.83&55.52&60.25&71.13&60.69&56.56&56.13 \\
				ICON &22.22&29.91&58.78&64.57&62.76&57.38&64.71&63.04&58.86&63.42&67.19&{\bf 60.81}&59.09&58.54\\
				\hline
				DialogueRNN &25.69&33.18&75.10&{\bf 78.80}&58.59&{\bf 59.21}&64.71&{\bf 65.28}&80.27&{\bf 71.86}&61.15&58.91&63.40&{\bf 62.75}\\
				\hline
			\end{tabular}
		}
		\caption{Comparison between DialogueRNN and baseline methods on IEMOCAP dataset; bold font denotes
			the best performances. Average(w) = Weighted average. ICON results differ from the original paper~\citep{hazarika2018icon} as in our experiment, we disregard their contextual feature extraction and pre-processing part. and  More details can be found in \citet{majumder2019dialoguernn}.}
		\label{tab:results-text1}
	\end{table*}
	
	\begin{table*}[t]
		\centering
		\begin{tabular}{c||c@{~~}c|c@{~~}c|c@{~~}c|c@{~~}c}
			\hline
			\multirow{3}{*}{Methods} & \multicolumn{8}{c}{SEMAINE}\\
			\cline{2-9} &\multicolumn{2}{c|}{Valence}& \multicolumn{2}{c|}{Arousal}& \multicolumn{2}{c|}{Expectancy} & \multicolumn{2}{c}{Power}\\
			\cline{2-9} & $MAE$ & $r$ & $MAE$ & $r$ & $MAE$ & $r$ & $MAE$ & $r$\\
			\hline
			\hline
			CNN &0.545&-0.01&0.542&0.01&0.605&-0.01&8.71&0.19 \\
			memnet &0.202&0.16&0.211&0.24&0.216&0.23&8.97&0.05\\
			bc-LSTM &0.194&0.14&0.212&0.23&0.201&0.25&8.90&-0.04 \\
			bc-LSTM+Att &0.189&0.16&0.213&0.25&0.190&0.24&8.67&0.10 \\
			\hline
			CMN &0.192&0.23&0.213&0.29&0.195&0.26&8.74&-0.02 \\
			ICON&0.181&0.245&0.19&0.317&0.185&0.272&8.45&0.244\\
			\hline
			DialogueRNN &{\bf 0.168}&{\bf 0.35}&{\bf 0.165}&{\bf 0.59}&{\bf 0.175}&{\bf 0.37}&{\bf 7.90}&{\bf 0.37}\\
			\hline
		\end{tabular}
		\caption{Comparison between DialogueRNN and baseline methods on SEMAINE dataset;Acc. = Accuracy,
			$MAE$ = Mean Absolute Error, $r$ = Pearson correlation coefficient; bold font denotes
			the best performances. More details can be found in \citet{majumder2019dialoguernn}.}
		\label{tab:results-text2}
	\end{table*}
	
\begin{figure}[!h]
		\centering
		\includegraphics[width=\linewidth]{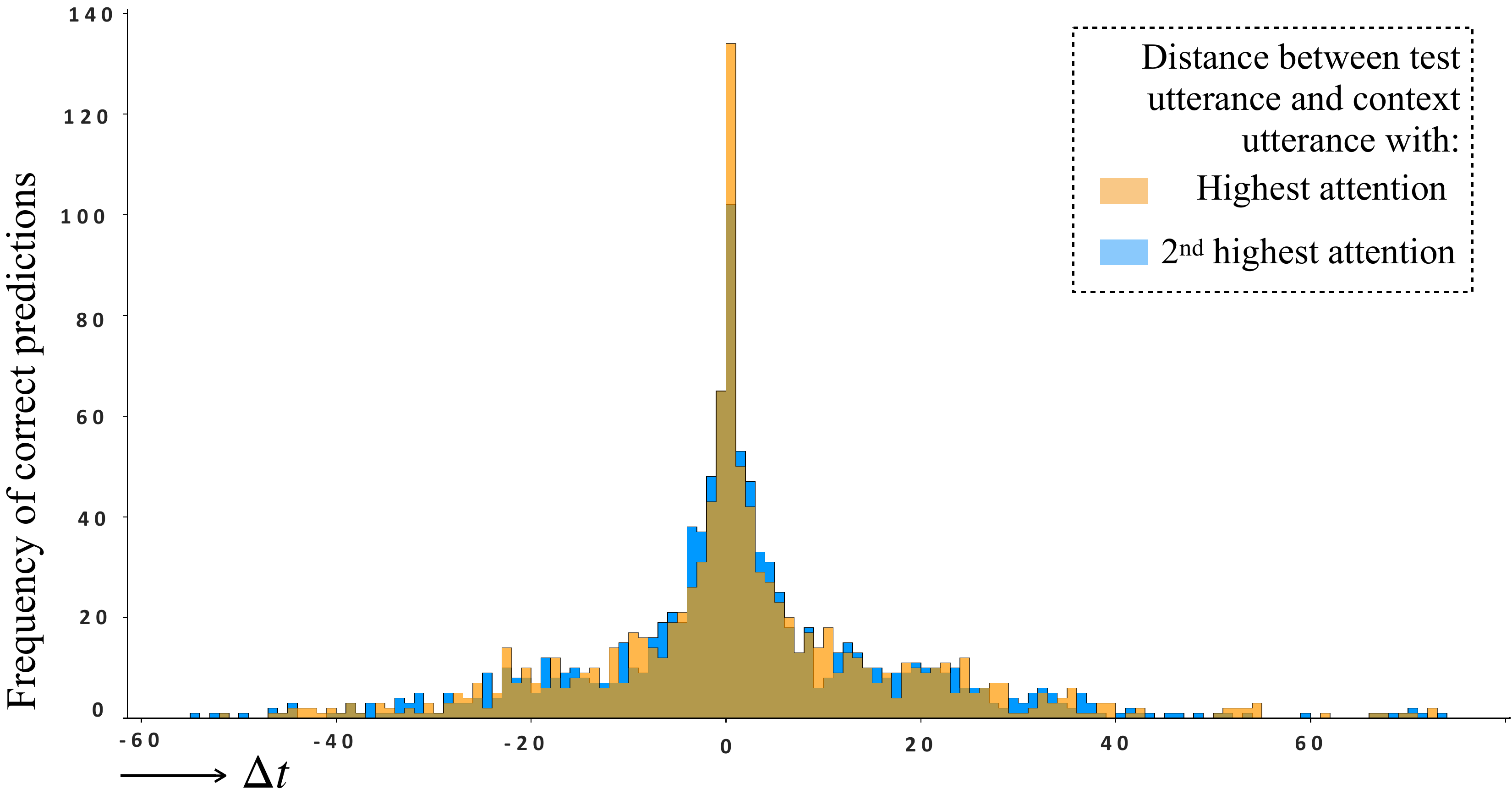} 
		\caption{Histogram of $\Delta t=$ distance between the target utterance and its context utterance based on DialogueRNN's attention scores. Note: Figure taken from \citet{majumder2019dialoguernn}.}
		\label{fig:DeltaTTrend}
	\end{figure}

	All of these models affirm that contextual history, modeling self and inter-speaker influence are beneficial to ERC (shown in \cref{fig:case-study} and \cref{fig:DeltaTTrend}). Further, DialogueRNN shows that the nearby utterances are generally more context rich and ERC performance improves when the future utterances, at time $>t$, are available. This is indicated by \cref{fig:DeltaTTrend}, where DialogueRNN uses both past and future utterances as context with roughly the same frequency. Also, the distant utterances are used less frequently than the nearby utterances. On the other hand, CMN and ICON do not use future utterances as context at all. However, for real-time applications, systems cannot rely on future utterances. In such cases, CMN, ICON and DialogueRNN with fixed context window would be befitting.
	
	All these networks, namely CMN, ICON, IANN and DialogueRNN, perform poorly on the utterances with emotion shift.
	In particular, the cases where the emotion of the target utterance differs from the previous utterance, DialogueRNN could only correctly predict $47.5\%$ instances. This stands less as compared to the $69.2\%$ success-rate that it achieves at the regions of no \emph{emotional-shift}. 
	
	Among these three approaches, only DialogueRNN is capable of handling multiparty conversations on large scale. However, on the multiparty conversational dataset MELD, only a little performance improvement (shown in \cref{tab:emotion_results}) is observed by DialogueRNN compared to bc-LSTM which depicts a future research direction on multiparty ERC. ICON and CMN are designed to detect emotions in dyadic dialogues. Adapting ICON and CMN to apply on multiparty conversational dataset MELD can cause scalability issue in situations when number speakers participating in a conversation in the test data is more than the training data.
	
	\begin{table*}[h]
		\centering
		\large
		\scalebox{0.9}{
			\begin{tabular}{c|cccccccc}
				\hline
				\multirow{2}{*}{Modality}&\multicolumn{8}{c}{{Emotions}}\\\cline{2-9}
				& anger & disgust & fear & joy & neutral & sadness & surprise& w-avg.\\\hline
				CNN & 34.49 & 8.22 & 3.74 & 49.39 & 74.88 & 21.05 & 45.45 & 55.02 \\
				bc-LSTM &42.06 &21.69 &7.75 &54.31 &71.63 &26.92 &48.15 & 56.44 \\
				DialogueRNN &40.59 &2.04 &8.93 &50.27 &75.75 &24.19 &49.38 &57.03 \\
				\hline
			\end{tabular}
		}
		\caption{Test-set F-score results of bc-LSTM and DialogueRNN for emotion classification in MELD. Note: \textit{w-avg} denotes weighted-average. text-CNN: CNN applied on text, contextual information were not used.}
		\label{tab:emotion_results}
	\end{table*}
	
	Due to the sequential nature of the utterances in conversations, RNNs are used for context generation in the aforementioned models. However, there is ample room for improvement, as the RNN-based context representation methods perform poorly in grasping long distant contextual information.
	
\begin{figure}[h]
		\centering
		\includegraphics[width=\linewidth]{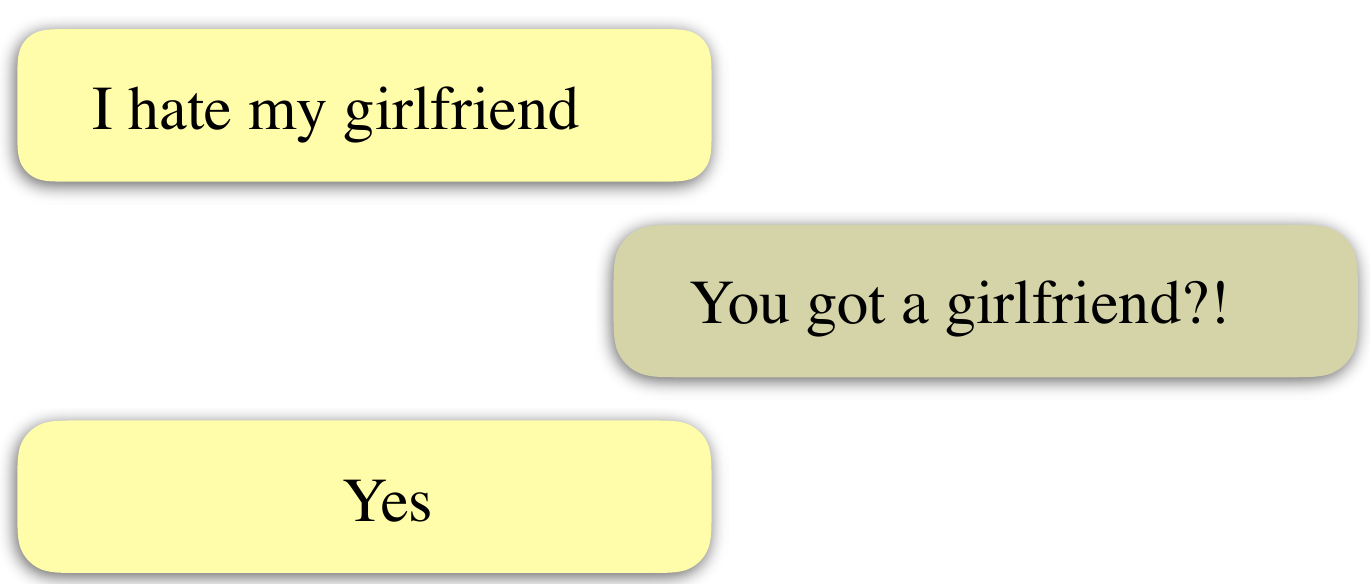}
		\caption{An example of a 3-turn conversation extracted from EmoContext dataset.}
		\label{fig:emocontext2}
	\end{figure}

	Recently, two shared tasks - EmotionX (co-located with SocialNLP workshop) and EmoContext~\footnote{https://www.humanizing-ai.com/emocontext.html} (co-located with Semeval 2019) have been organized to address the ERC problem. 
	EmoContext shared task has garnered more than 500 participants, affirming the growing popularity of this research field. Compared to other datasets, EmoContext dataset~\citep{chatterjee2019understanding} has very short conversations consisting only three utterances where the goal is to label the 3rd utterance as shown in \cref{fig:emocontext2}.
	Emotion labels of the previous utterances are not present in the EmoContext dataset. The key works~\citep{bae2019snu_ids, huang2019ana, chatterjee2019understanding} on this dataset have mainly leveraged on context modeling using bc-LSTM architecture~\citep{poria2017context} that encapsulates the temporal order of the utterances using an LSTM. A common trend can be noticed in these works, where traditional word embeddings, such as Glove~\citep{pennington2014glove}, are combined with contextualized word embeddings, such as ELMo~\citep{peters2018deep} to improve the performance.
	\begin{figure}[!h]
		\centering
		\includegraphics[width=\linewidth]{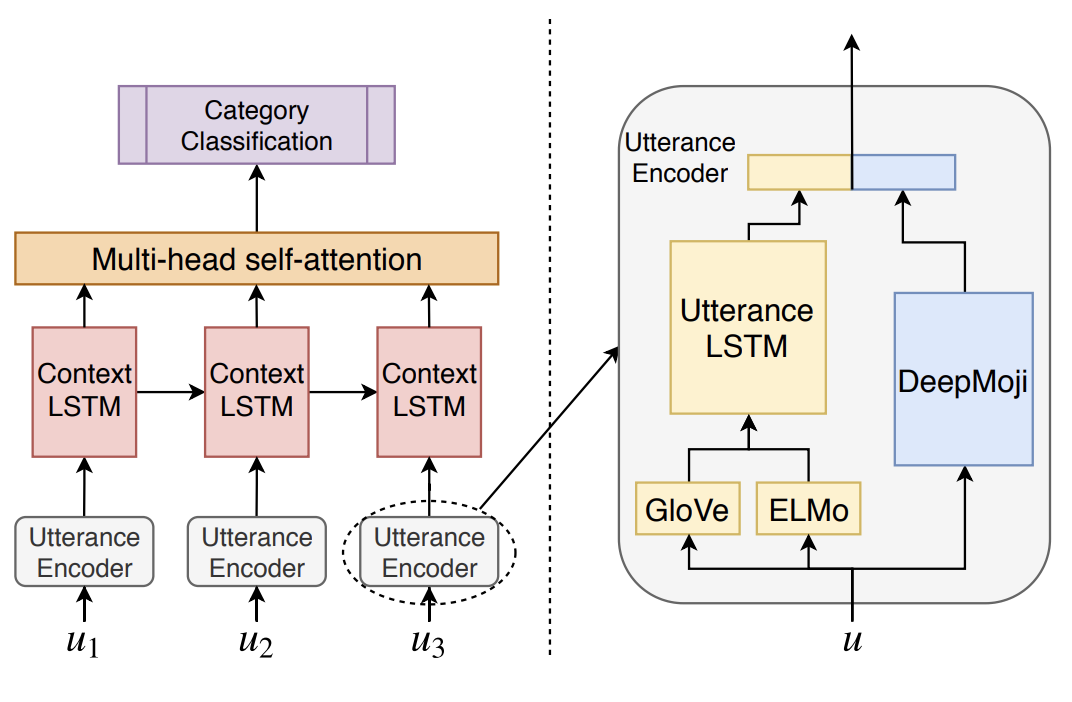}
		\caption{HRLCE framework applied on the EmoContext dataset.}
		\label{fig:emocontext}
	\end{figure}
	In \cref{fig:emocontext}, we depict the HRLCE framework, proposed by \citet{huang2019ana}, that comprises of an utterance encoder and a context encoder that takes input from the utterance encoder. To represent each utterance, HRLCE utilizes ELMo~\citep{peters2018deep}, Glove~\citep{pennington2014glove} and Deepmoji~\citep{felbo2017using}. 
	\begin{table}[!h]
		\centering
		\begin{tabular}{|c|c|}
			\hline
			Framework & F1 \\
			\hline
			HRLCE&76.66\\
			DialogueRNN&75.80\\
			\hline
		\end{tabular}
		\caption{HRLCE and DialogueRNN on the EmoContext dataset.}
		\label{tab:emocontextexp}
	\end{table}
	The context encoder in HRLCE adapts the bc-LSTM framework followed by a multi-head attention layer. \citet{huang2019ana} applied HRLCE framework only on the EmoContext dataset. However, HRLCE can be easily adapted to apply on other ERC datasets. It should be noted that none of the works on the EmoContext dataset utilize speaker information. In fact, in our experiments, we found that DialogueRNN, which makes use of the speaker information, performs similar (\cref{tab:emocontextexp}) to \citet{bae2019snu_ids, huang2019ana} and \citet{chatterjee2019understanding} on EmoContext dataset. One possible reason for this could be the presence of very short context history in the dataset that renders speaker information inconsequential.
	
	\section{Conclusion}
	\label{sec:conclusion}
	Emotion recognition in conversation has been gaining popularity among NLP researchers.
	In this work, we summarize the recent advances in ERC and highlight several key research challenges associated with this research area.
	Further, we
	point out how current work has partly addressed these challenges, while also presenting some shortcomings. Overall, we surmise that an effective emotion-shift recognition model and context encoder can yield
	significant performance improvement over chit-chat dialogue, and even improve some aspects of task-oriented dialogue. Moreover,  challenges like topic-level speaker-specific emotion recognition, ERC on multiparty conversations, and conversational sarcasm detection
	can form new research directions. Additionally, fine-grained speaker-specific continuous emotion
	recognition may become of interest for the purpose of tracking emotions during long monologue. We believe that addressing each of the challenges outlined in this paper will not only enhance AI-enabled conversation understanding, but also improve the performance of dialogue systems by catering to affective information.
	
	\bibliographystyle{IEEEtranN}
	\bibliography{refs}
\end{document}